# Non-Intrusive Electrical Appliances Monitoring and Classification using K-Nearest Neighbors

Mohammad Mahmudur Rahman Khan[1], Md. Abu Bakr Siddique[2], Shadman Sakib[3]
[1]Department of ECE, Vanderbilt University, Nashville, TN-37235, USA
[2,3]Department of EEE, International University of Business Agriculture and Technology, Dhaka-1230, Bangladesh
mohammad.mahmudur.rahman.khan@vanderbilt.edu[1], absiddique@iubat.edu[2], bbkrsddque@gmail.com[2], sakibshadman15@gmail.com[3]

*Abstract*—Non-Intrusive Load Monitoring (NILM) is the method of detecting an individual device's energy signal from an aggregated energy consumption signature [1]. As existing energy meters provide very little to no information regarding the energy consumptions of individual appliances apart from the aggregated power rating, the spotting of individual appliances' energy usages by NILM will not only provide consumers the feedback of appliance-specific energy usage but also lead to the changes of their consumption behavior which facilitate energy conservation. B Neenan et al. [2] have demonstrated that direct individual appliance-specific energy usage signals lead to consumers' behavioral changes which improves energy efficiency by as much as 15%. Upon disaggregation of an energy signal, the signal needs to be classified according to the appropriate appliance. Hence, the goal of this paper is to disaggregate total energy consumption data to individual appliance signature and then classify appliance-specific energy loads using a prominent supervised classification method known as K-Nearest Neighbors (KNN). To perform this operation we have used a publicly accessible dataset of power signals from several houses known as the REDD dataset. Before applying KNN, data is preprocessed for each device. Then KNN is applied to check whether their energy consumption signature is separable or not. KNN is applied with K=5.

## I. Introduction

The deficiency of reliable and sustainable energy sources is one of the greatest challenges the world is going to face in the very near future. The development of next-generation power systems, known as the smart grid, aims to bring computational intelligence and data mining technologies into the physical systems including households to shrink energy consumption. Recent studies [3] have revealed that energy disaggregation which separates an aggregated energy signal into the consumption of item-wise, appliance-by-appliance energy in a household can significantly encourage users to consume a smaller amount of energy and guide them how to act so.

Though some modern appliances are capable of communicating with utilities and item-wise energy bills can be automatically generated, older appliances do not have that ability. An alternative manner is to install additional sensors for each appliance to measure their individual energy consumption. However, this process is intrusive and sometimes is not a cost-effective option. Another measure that can be taken is the disaggregation of the whole electricity load into separate appliance loads using machine learning and signal processing methods [4]. This problem is referred to as Non-Intrusive Load Monitoring (NILM) which is the focus of this paper.

Several parametric and non-parametric classification methods have been employed to determine the type of appliance that the energy signal belongs to. Parametric classification methods usually presume that the data follows a probability distribution with some unknown model parameters which need to be estimated from the training data. For many real-life applications, this strong assumption may not hold and the prediction performance is reduced. Unlike parametric methods, non-parametric NILM methods do not make such assumptions and learn the predictive function directly from the data. In literature, the nearest-neighbor based method, such as K nearest neighbors (KNN) gains its popularity due to its simplicity and effectiveness. Though recent studies [5] have shown that KNN suffers from the learning performance reduction for the data that is class-distribution imbalanced, in case of energy data disaggregation the KNN demonstrated good performance.

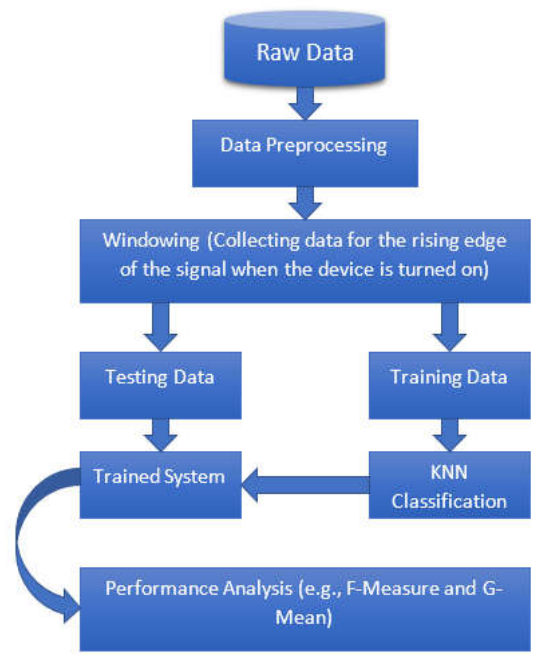

Fig. 1. Block diagram representing the whole process taken in this paper for electrical appliances classification by KNN step by step

In this paper, we implement the KNN algorithm for energy data disaggregation. We employ the KNN algorithm for identifying the devices for a given power consumption signal and evaluate its performance over a real-life data set. Three performance measures including overall accuracy, G-mean and F-measure are used for performance evaluation.

Experimental results show that the KNN performs reasonably well in case of classifying the energy consumption signal signature of devices.

## II. Literature Review

These days, the conservation of energy is a difficult issue caused by exponentially growing energy demands. To discover this issue experts are endeavoring to propagate new technological arrangements everywhere all over the world. Hart [1] describes a procedure where energy disaggregation correspondingly denoted as a non-intrusive load monitoring (NILM) intends to separate a home's total power consumption into specific appliances. It is an advantageous method for deciding the energy consumption and the condition of activity of individual appliances dependent on the investigation of the aggregate load estimated by the major power meter in an organization. This technique was previously developed to unravel the energy use data accumulation by the utilities. As an augmentation to Hart's effort, Ducange et al. [6] propounded a twined load distinguishing proof algorithm, in which a blending of finite-state machine along with fuzzy transitions method was entreated to recognize appliance. A novel strategy proposed by Parson et al. [7] utilizing Hidden Markov Model (HMM) algorithm for Non-intrusive Appliance Monitoring (NIALM) where separate appliances are iteratively isolated from the cumulative load. HMMs have been occupied to characterize the device conduct also [8]. A cogent confabulation (CC) neural network algorithm [9] introduced by Park et al. for the appliance recognition with the utilization of a non-intrusive home energy screen. In their research, the achievement rate was 83.8% which is superior to the artificial neural network draws strategies. In another investigation, NILM dependent on Long Short-Term Memory Recurrent Neural Network (LSTM-RNN) model with deep learning and novel signature proposed by Kim et al. [10]. For their investigation, they have utilized the REDD data set just as the UK-DALE data set. Thus, by associating at their trial results they have demonstrated that their method is more powerful than the existing NILM method. Both supervised and unsupervised machine learning strategies have been used in the quest for an optimized NILM classification. Supervised learning requires different isolated training samples for every appliance, a monotonous task most consumers are reluctant to perform. Conversely, unsupervised learning methods can disaggregate appliances without training samples; however, their capacity to determine appliances is constrained since they can just find appliances that happen to ensue in isolation and require pre-existing appliance models. Moreover, for electricity disaggregation supervised learning models' applications comprise Semi-Definite Program Relaxation with Randomized Rounding [11], and Markov random fields [12], etc. Though these supervised learning methods have constrained application in real-world settings since they require labeled samples for training or separate meters for every appliance, the two of which are costly to get. An unsupervised NILM method proposed utilizing another openly accessible data set BLUED for the easier establishment process, more extensive materialness in the private division, and less intrusion in the monitoring procedure.

Despite huge research in electricity disaggregation, existing methods remain unimplemented in reality. In this unsupervised method of NILM, high event discovery inclusive of clustering acquired with disaggregation up to 92% of the aggregate, consumed energy [13]. In another investigation, Convolutional Neural Network (CNN) is applied to NILM for the detection of residential apparatus [14]. CNN network was developed to distinguish the kind of apparatus from the transient power signal data acquired right at the instant an apparatus is connected. While the NIALM issue has gotten consideration since the mid-1990s, there is still yet a lot to be made in the field of energy signature disaggregation to carry it to the business front.

## III. Description of K-Nearest Neighbors (KNN) Classifier

K-nearest neighbor (KNN) [15] is a nonparametric and case based idle learning classifier used to forecast the categorization of a new test point in a dataset where data points are divided into a number of classes. Since KNN preserves all the existing instances and necessitates scrutinizing entire dataset to categorize a new test point, the nominal training but large-scale testing period of KNN happens equally at memory and computational expenses. KNN is a supervised learning composed of a specified labeled dataset accommodating training sets (x, y) and would like to represent the correlation between x and y. The objective of KNN is to find out a function $h: x \rightarrow y$ hence for a new test point x, h(x) can assertively deduce the equivalent output y. In KNN categorization, a new test point is categorized by greater number of votes of its neighbors, with the test point being allocated to the category most available among its k nearest neighbors. If k=1, then new point is allocated to the category of its only nearest neighbor. Figure 1 illustrates the fundamental diagram of KNN classifier.

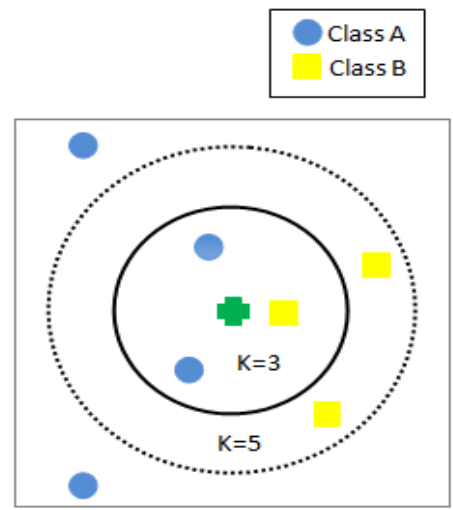

Fig. 2. Fundamental design of KNN classifier. If k = 3 the test sample (green polygon) is allocated to the category of blue circle and if k = 5 then it is assigned to the category of yellow square

The feat of KNN classifier is mostly hinges on the distance metric employed to recognize the k nearest neighbors of a test point. The most routinely employed one is the Euclidean distance represented by:

$$d(x,x') = \sqrt{(x_1 - x'_1)^2 + (x_2 - x'_2)^2 + \ldots + (x_n - x'_n)^2} \ \ldots\ (1)$$

For a specified number of nearest neighbors, k, and an unidentified test point, x, and a distance metric d, a KNN classifier executes the two steps as following:
1. It explores throughout the whole dataset and calculates d between x and every training assessment. Let presume that the k points in the training data that are adjoining to x be in set U. It is important to note that k typically to be an odd number as it restrains tied condition.
2. Then it evaluates conditional probability for every category i.e. the probability for a portion of points exists in the set U for a certain category. In conclusion, the unidentified test point x is allocated to the category with maximum probability.

$$P(y = j | X = x) = \frac{1}{K} \sum_{i \in U} I\left(y^{(i)} = j\right) \ \ldots\ (2)$$

IV. OVERVIEW OF DATASET

A. REDD Dataset

Reference Energy Disaggregation Data Set (REDD) [16], is the first openly accessible leading open data set made specifically for disaggregation where true loads of each house is recognized. The dataset comprises equally aggregate and sub-metered power data from six distinct houses in Massachusetts of the United States and has since turned into the most famous data set for assessing energy disaggregation algorithms. The data is explicitly equipped towards the conduction of energy disaggregation that indicates the component devices from an aggregated electricity signal. Besides, it gives a powerful depiction corresponding to the effects happened regarding the energy in the house over the period. From around forty homes in the Boston and San Francisco urban regions the data were collected. Among them, the mainstream of the data was assembled for 48 distinctive circuit breakers, with the accumulation period for each home ordinarily lengthening from two weeks to one month. The entire home voltage and current in each home, at high frequencies, (16 kHz) were checked by the analysts, to trace the genuine AC waveforms of the total electrical energy signatures in the houses. Circuits are named with understandable depictions, with a portion of the significant loads present on the circuits recorded. Throughout the experiment, the appliance being observed is marked in the data set. The entire home signal gives high recurrence data where devices can be distinguished by the waveforms, while the per-circuit and per-plug signals give the ground truth regarding what occurred in the home.

B. Preprocessed REDD Dataset

From the REDD dataset, we picked and preprocessed energy consumption signals for seven unique devices from the five houses for appliances' classification purposes, using KNN, which gave a decent test environment to our calculation. These seven devices are Furnace, Bath GFI, Oven, Electronics, Kitchen Outlet, Washer Dryer, and Microwave-oven. We have taken each device's energy consumption signal, cropped where there is any kind of activity in the signal and made the dataset by accumulating those cropped points. Thus data is accumulated for seven different devices from their corresponding energy consumption signals. The sample raw and processed energy consumption data for Electronics and Microwave-oven are shown in figure 3 and figure 4 respectively.

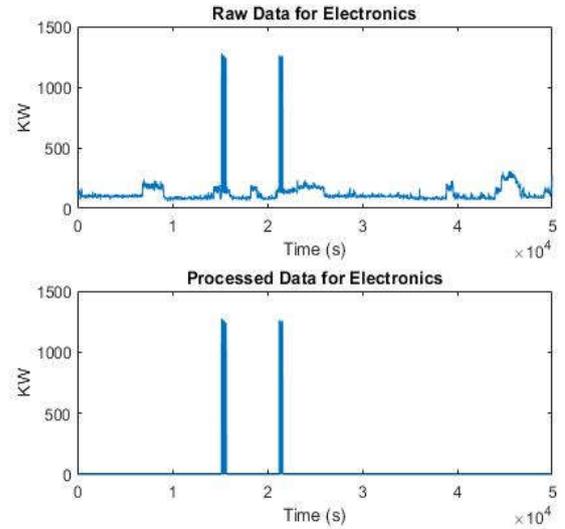

Fig. 3. Raw and processed energy consumption data for Electronics

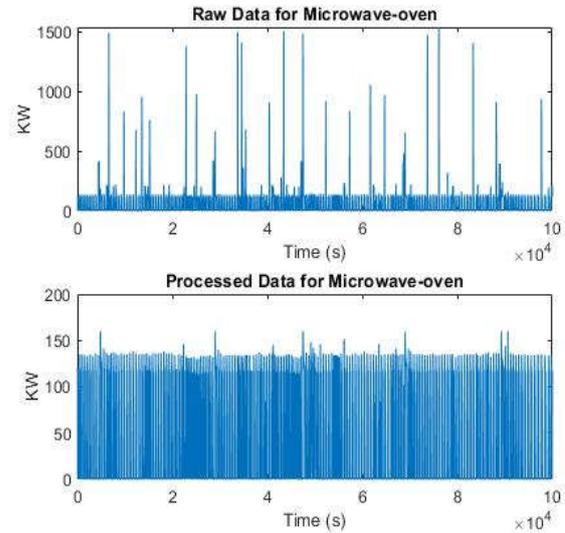

Fig. 4. Raw and Processed energy consumption data for Microwave-oven

## V. RESULTS AND DISCUSSION

Though the accuracy metric is widely used for evaluating the classification performance, it may lead to a deceiving conclusion particularly for imbalanced data set where the minority is usually underrepresented and it becomes a biased performance measure towards the majority class [17]. Hence we are considering F-measure and G-mean as performance evaluation metrics for the classifier.

### A. Performance Evaluation Metrics

Let's consider a two-class problem where TP, TN represents true positive and true negative labels. The overall performance of the classifier can be clearly viewed as a confusion matrix, shown in Table 1.

TABLE I. CONFUSION MATRIX

|  | **Predicted Positive** | **Predicted Negative** |
|---|---|---|
| Actual Positive | True Positives (TP) | False Negatives (FN) |
| Actual Negative | False Positives (FP) | True Negatives (TN) |

In Table 1, TP (true positive) represents the bunch of positive samples that are identified appropriately, TN (true negative) indicates the bunch of negative samples that are identified appropriately. On the other hand, FP (false positive) is the bunch of the positive points that are misidentified as negative and FN (false negative) represents the bunch of negative points that are misidentified as positive. Given the confusion matrix, the overall accuracy or error rate can be

$$Accuracy = \frac{TP + TN}{TP + FP + TN + FN} \ldots\ldots(3)$$

$$Error\ Rate = 1 - Accuracy \ldots\ldots(4)$$

For better performance evaluation, we consider two comprehensive performance metrics, F-measure and G-Mean, which are expressed as follows:

$$F - measure = \frac{2 \times Precision \times Recall}{Precision + Recall} \ldots\ldots(5)$$

$$G - mean = \sqrt{Precision + Recall} \ldots\ldots(6)$$

where the Precision and Recall can be evaluated from the confusion matrix which are defined:

$$Precision = \frac{TP}{TP + FP} \ldots\ldots(7)$$

$$Recall = \frac{TP}{TP + FN} \ldots\ldots(8)$$

Since the precision value symbolizes the exactness of the classifier and the recall value indicates the completeness of the classifier [18], the F-measure and G-mean which are the combination of both precision and recall are more comprehensive and widely used in imbalanced learning.

### B. Result Analysis

For evaluating the performance of the KNN, the whole data set is split into two parts: training (90% of the data) and testing (10% of the data). Table 2 presents the F-measure and G-mean values based on which the performance of KNN could be analyzed.

TABLE II. CHANNEL-WISE AND OVERALL F-MEASURE AND G-MEAN VALUES FOR KNN CLASSIFIER

| **Channels** | **F-measure** | **G-mean** |
|---|---|---|
| Furnace | 1.00 | 1.00 |
| Bath GFI | 0.917 | 0.917 |
| Oven | 0.900 | 0.900 |
| Electronics | 0.667 | 0.707 |
| Kitchen Outlet | 1.00 | 1.00 |
| Washer Dryer | 1.00 | 1.00 |
| Microwave Oven | 1.00 | 1.00 |
| Overall | 0.926 | 0.932 |

It is clearly visible from Table 2 that the KNN algorithm is very efficient in identifying the Furnace, Kitchen outlet, Washer dryer, and the Microwave oven energy signals. For Electronic devices, it demonstrated the least efficiency. Nevertheless, it is evident that the KNN method can perform reasonably well in the case of energy data classification. The following confusion matrixes show the overall classification performance.

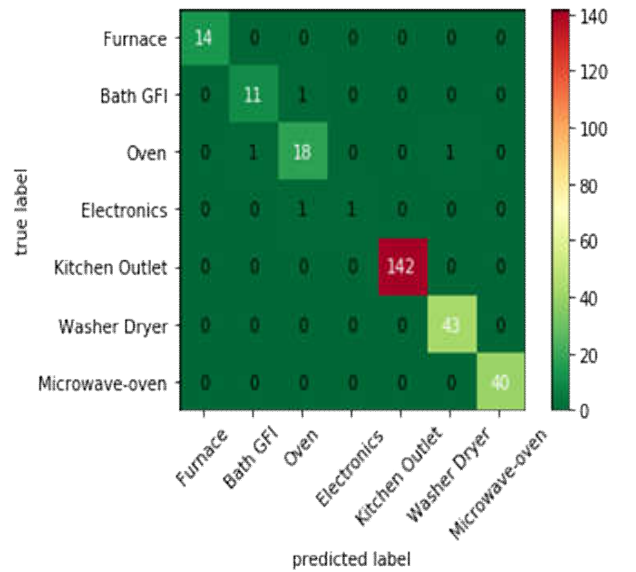

Fig. 5. Confusion matrix for energy consumption signal classification - actual

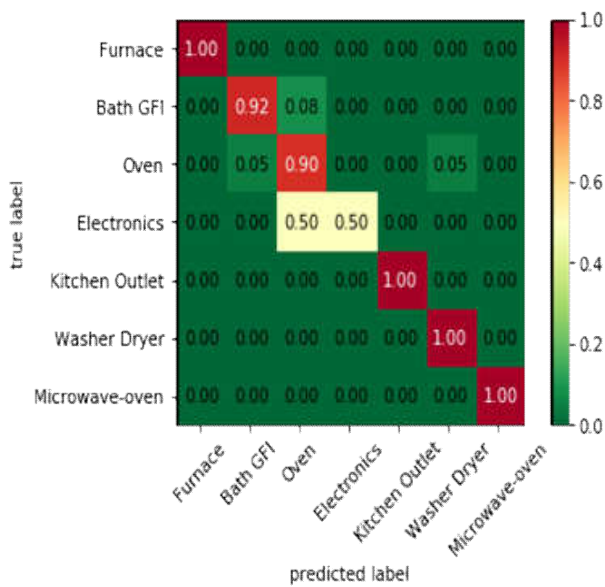

Fig. 6. Confusion matrix for energy consumption signal classification – normalized

The KNN could classify almost all the energy data here. However, it got confused while classifying the electronics devices signal. On the contrary, for the other channels (i.e., Furnace, Kitchen outlet, Washer dryer, and Microwave oven) the performance of the classifier was significantly good.

## VI. CONCLUSION

In summary, in this paper, a general study is conducted on household power consumption data from the REDD dataset to classify and identify seven different electrical appliances by their unique energy consumption signatures using K Nearest Neighbors (KNN) classification algorithms. The training and testing data ratio is maintained as 90%: 10% throughout the whole process. It is evident from the result and discussion section that the KNN classifier is very effective in identifying appliances from energy consumption signals and can capture the nonlinearities of the signals efficiently. Though for electronic devices KNN classifier got confused and gave the lowest classification accuracy of 50%, KNN is very effective in classifying Furnace, Kitchen outlet, Washer dryer and Microwave oven with an accuracy of 100% for each device.

REFERENCES

[1] G. W. Hart, "Nonintrusive appliance load monitoring," *Proceedings of the IEEE,* vol. 80, pp. 1870-1891, 1992.
[2] B. Neenan*, et al.*, "Residential electricity use feedback: A research synthesis and economic framework," *Electric Power Research Institute,* vol. 3, 2009.
[3] C. Fischer, "Feedback on household electricity consumption: a tool for saving energy?," *Energy efficiency,* vol. 1, pp. 79-104, 2008.
[4] S. M. Tabatabaei*, et al.*, "Toward non-intrusive load monitoring via multi-label classification," *IEEE Transactions on Smart Grid,* vol. 8, pp. 26-40, 2016.
[5] J. Z. Kolter and T. Jaakkola, "Approximate inference in additive factorial hmms with application to energy disaggregation," in *Artificial Intelligence and Statistics,* 2012, pp. 1472-1482.
[6] P. Ducange*, et al.*, "A novel approach based on finite-state machines with fuzzy transitions for nonintrusive home appliance monitoring," *IEEE Transactions on Industrial Informatics,* vol. 10, pp. 1185-1197, 2014.
[7] O. Parson*, et al.*, "Using hidden markov models for iterative non-intrusive appliance monitoring," 2011.
[8] O. Parson*, et al.*, "Non-intrusive load monitoring using prior models of general appliance types," in *Twenty-Sixth AAAI Conference on Artificial Intelligence*, 2012.
[9] S. W. Park*, et al.*, "Appliance identification algorithm for a non-intrusive home energy monitor using cogent confabulation," *IEEE Transactions on Smart Grid,* vol. 10, pp. 714-721, 2017.
[10] J. Kim*, et al.*, "Nonintrusive load monitoring based on advanced deep learning and novel signature," *Computational intelligence and neuroscience,* vol. 2017, 2017.
[11] K. Shaloudegi*, et al.*, "SDP relaxation with randomized rounding for energy disaggregation," in *Advances in Neural Information Processing Systems*, 2016, pp. 4978-4986.
[12] S. Tomkins*, et al.*, "Disambiguating Energy Disaggregation: A Collective Probabilistic Approach," in *IJCAI*, 2017, pp. 2857-2863.
[13] K. S. Barsim*, et al.*, "Unsupervised Non-Intrusive Load Monitoring of Residential Appliances."
[14] D. de Paiva Penha and A. R. G. Castro, "Convolutional neural network applied to the identification of residential equipment in non-intrusive load monitoring systems," in *3rd International Conference on Artificial Intelligence and Applications*, 2017, pp. 11-21.
[15] T. Cover and P. Hart, "Nearest neighbor pattern classification," *IEEE transactions on information theory,* vol. 13, pp. 21-27, 1967.
[16] J. Z. Kolter and M. J. Johnson, "REDD: A public data set for energy disaggregation research," in *Workshop on Data Mining Applications in Sustainability (SIGKDD), San Diego, CA*, 2011, pp. 59-62.
[17] N. C. Truong*, et al.*, "Forecasting multi-appliance usage for smart home energy management," in *Twenty-Third International Joint Conference on Artificial Intelligence*, 2013.
[18] S. Ravindran, "Learning Rare Patterns with Multilevel Models," 2017.